\documentclass[a4paper,fleqn]{cas-sc}

\usepackage{makecell}

\usepackage{soul}
\usepackage{color}
\usepackage{xcolor}
\usepackage[table]{xcolor}
\soulregister{\cite}7 
\soulregister{\citep}7 
\soulregister{\citet}7 
\soulregister{\ref}7 
\soulregister{\pageref}7 

\def\tsc#1{\csdef{#1}{\textsc{\lowercase{#1}}\xspace}}
\tsc{WGM}
\tsc{QE}

\begin{document}
\let\WriteBookmarks\relax
\def\floatpagepagefraction{1}
\def\textpagefraction{.001}

\shorttitle{Real-time OmniMVS for Real Scenarios based on Teacher-Student Learning with Unlabeled Data}

\shortauthors{M. Li et~al.}

\title [mode = title]{Real-time Multi-view Omnidirectional Depth Estimation for Real Scenarios based on Teacher-Student Learning with Unlabeled Data}



%

\author[1]{Ming Li}[orcid=0000-0002-1341-5585]



\ead{mingli@nuist.edu.cn}


\credit{Conceptualization, Methodology, Software, Validation, Writing - Original Draft, Writing - Review and Editing, Funding acquisition}

\affiliation[1]{organization={School of Artificial Intelligence/School of Future Technology, Nanjing University of Information Science and Technology},
  city={Nanjing},
  postcode={210044},
  country={China}}

\author[2]{Xiong Yang}
\credit{Software, Data Curation, Visualization}
\author[2]{Chaofan Wu}
\credit{Software, Data Curation}
\author[2]{Jiaheng Li}
\credit{Data Curation, Visualization}
\author[2]{Pinzhi Wang}
\credit{Data Curation}

\affiliation[2]{organization={School of Electronic Science and Engineering, Nanjing University},
  city={Nanjing},
  postcode={210023},
  country={China}}

\author[3]{Xuejiao Hu}
\credit{Validation}
\author[2]{Sidan Du}
\ead{coff128@nju.edu.cn}
\credit{Supervision, Project administration}

\author[2,4]{Yang Li}
\cormark[1]
\ead{yogo@nju.edu.cn}
\credit{Supervision, Funding acquisition}




\affiliation[3]{organization={School of Computer Engineering, Jinling Institute of Technology},
  city={Nanjing},
  postcode={211169},
  country={China}}

\affiliation[4]{organization={Suzhou High Technology Research Institute, Nanjing University},
  city={Suzhou},
  postcode={215123},
  country={China}}

\cortext[1]{Corresponding author}



\begin{abstract}
  Omnidirectional depth estimation enables efficient 3D perception over a full 360-degree range. However, in real-world applications such as autonomous driving and robotics, achieving real-time performance and robust cross-scene generalization remains a significant challenge for existing algorithms. In this paper, we propose a real-time omnidirectional depth estimation method for edge computing platforms named Rt-OmniMVS, which introduces the Combined Spherical Sweeping method and implements the lightweight network structure to achieve real-time performance on edge computing platforms. To achieve high accuracy, robustness, and generalization in real-world environments, we introduce a teacher-student learning strategy. We leverage the high-precision stereo matching method as the teacher model to predict pseudo labels for unlabeled real-world data, and utilize data and model augmentation techniques for training to enhance performance of the student model Rt-OmniMVS. We also propose HexaMODE, an omnidirectional depth sensing system based on multi-view fisheye cameras and edge computation device. A large-scale hybrid dataset contains both unlabeled real-world data and synthetic data is collected for model training. Experiments on public datasets demonstrate that proposed method achieves results comparable to state-of-the-art approaches while consuming significantly less resource. The proposed system and algorithm also demonstrate high accuracy in various complex real-world scenarios, both indoors and outdoors, achieving an inference speed of 15 frames per second on edge computing platforms.
\end{abstract}


\begin{highlights}
  \item Reducing the amount of interpolation can enhance the efficiency on edge devices
  \item Pseudo-labels can effectively facilitate the training on real-world unlabeled data
  \item Data and model augmentations can significantly improve the model performance
\end{highlights}


\maketitle


\section{Introduction}
Depth estimation of the environment is the foundation of autonomous driving and obstacle avoiding of robots. Visual depth estimation with camera sensor offers advantages such as low cost, high density, and rich semantic information. Recently, omnidirectional depth estimation has attracted the attention of researchers because of its efficiency in perceiving the surrounding 3D environment. Some of the omnidirectional depth estimation methods use single \cite{3D60,Wang2020bifuse,Jiang2021unifuse} or multiple \cite{360SD-Net,Li2021csdnet,Li2022mode,Li2024robustmode} panoramas as inputs to predict the corresponding depth map. For more widespread practical applications, many algorithms utilize multiple fisheye cameras arranged in a surround view configuration to achieve 360$^\circ$ coverage and acquire depth information \cite{SweepNet, Omnimvs,Omnimvs_J,Wang2024Casomnimvs,Jiang2024Romnistereo}. Real-world applications require high-speed enviroment perception on embedded edge computing platforms with limited computational power and energy consumption. However, only a few methods \cite{Jiang2024Romnistereo,Deng2025OmniStereo,realtimeOmni,Wang2023FastOmniMVS} targeting optimization for real-time performance. Furthermore, achieving high accuracy and robustness in diverse and complex environments is also an urgent requirement for real-world applications. However, most existing methods are trained and validated on synthetic datasets, resulting in limited cross-domain generalization ability. Due to the high cost of acquiring dense 360$^\circ$ depth labels in real-world scenarios, training models using unlabeled real-world data remains a major challenge in omnidirectional depth estimation. A few existing methods \cite{leeSemiSupervised2022, Chen2023UnOmnimvs} leverage image reprojection for unsupervised learning, but the performance remain insufficient to meet practical demands.

Our previous work CasOmniMVS \cite{Wang2024Casomnimvs} proposes a multi-stage cascaded network architecture with dynamic adjustment of the hypothetical spherical range. This design effectively reduces mismatches of multi-view features to achieve high accuracy, but still fails to enable real-time inference on edge devices. Although some approaches \cite{Jiang2024Romnistereo,Deng2025OmniStereo} achieve fast speed on GPU platforms, it's still lack of optimization for edge computing platforms. The construction of spherical cost volume requires numerous operations such as random memory access and memory copy, which has a significant impact on the inference speed of edge devices. Due to the computational amount of projection interpolation operations and complex operators such as 3D convolutions, real-time inference on edge computing platforms remains a significant challenge. In summary, real-time performance on edge computing platforms and generalization ability cross diverse scenarios are the primary challenges faced in the field of multi-view omnidirectional depth estimation.

To achieve high-accuracy and real-time omnidirectional depth estimation in complex and diverse real-world scenarios, this paper proposes an optimized lightweight network architecture and introduces a teacher-student training strategy to enable training with unlabeled real-scenarios data. Specifically, to enhance computational efficiency on edge computation devices, we introduce a Combined Spherical Sweeping method to reduce the amount of interpolation operations in matching cost building, and propose a real-time algorithm named Rt-OmniMVS with a 2D-CNN based lightweight cost aggregation network. To achieve high accuracy and robustness for proposed Rt-OmniMVS model in complex real-world scenarios, we introduce a teacher-student training strategy, which uses teacher model to generate pseudo-lables for unlabeled real data and trains the real-time student model with techniques such as data augmentation and model augmentation. Besides, the proposed algorithm decouples the camera number and layout from the model structure, leading to the improvement of generalization capability and flexibility across different camera setups.

\begin{figure}[]%
  \centering
  \includegraphics[width=0.5\textwidth]{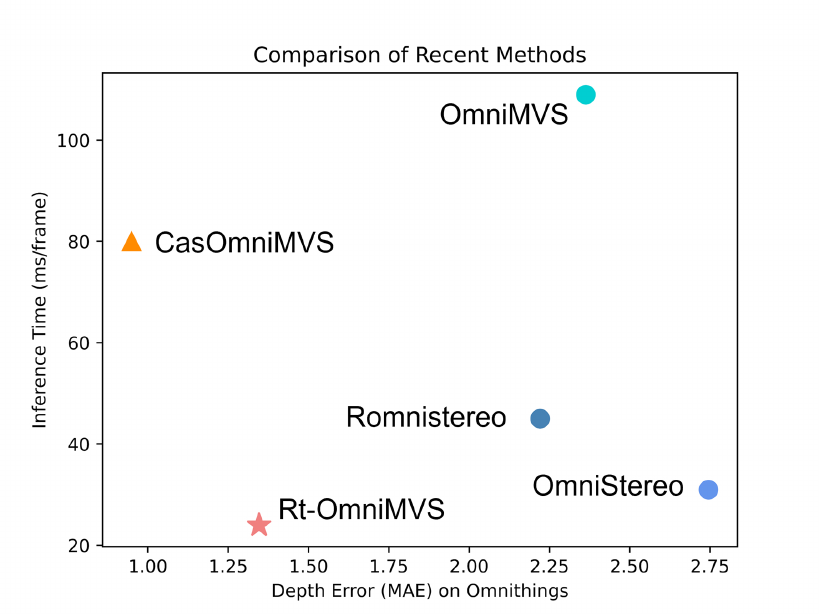}
  \caption{The comparison of recent multi-view omnidirectional depth estimation methods on accuracy and inference time. The proposed Rt-OmniMVS has achieved competitive accuracy performance with fastest inference speed}\label{fig:combined_sweeping}
\end{figure}

We also propose a Multi-view Omnidirectional Depth Estimation (MODE) system to collecte large-scale real-scene data and validate the algorithm on diverse scenarios. The proposed system is built on a robot chassis. To allow larger overlapping regions for multiple fisheye cameras, we utilize a hexagonal arrangement of six fisheye cameras. The proposed HexaMODE (\textbf{hexa}gonal MODE) System adopts an edge computing platform, NVIDIA Jeston Orin for model inference and system control. Furthermore, we construct a hybrid dataset Hexa360Depth that comprises real-world and synthetic scenes for model training. The proposed Rt-OmniMVS achieves a inference speed of more than 15 frames per second (fps) on the NVIDIA Orin platform on HexaMODE system, demonstrating high accuracy, robustness, and generalization performance in real-world scenarios. More details and experiments can be found at \url{https://nju-ee.github.io/Autonomous_Driving_Research_Group.page/depth/}

In summary, the main contributions of this work are as follows:

\begin{itemize}
  \item We introduce a combined spherical sweeping algorithm and develop a lightweight network Rt-OmniMVS for multi-view omnidirectional depth estimation tailored to edge devices, which substantially reduces computation complexity, achieving an inference speed of 15 fps on NVIDIA AGX Orin, and also presents the generalization capability cross various camera settings. Experimental results on public datasets demonstrate that the proposed method over-performs existing approaches in inference speed with competitive accuracy performance.
  \item We propose a teacher-student learning strategy with data and model augmentation techniques to train the Rt-OmniMVS model effectively with unlabeled real-world data, enables cross-task knowledge transfer. The proposed approach yields high accuracy, robustness, and generalization in real-world applications.
  \item We present HexaMODE, an omnidirectional depth estimation system utilizing a six-fisheye camera configuration and an edge computing platform, to collect real-world data and validate the algorithm in complex scenarios. We also propose Hexa360Depth, a large-scale hybrid dataset consists of real-world and synthetic scenes, which contains scenarios with diverse environment conditions and depth distributions.
\end{itemize}

\section{Related work}
\subsection{Omnidirectional Depth Estimation}
\textbf{Monocular omnidirectional depth estimation.} Zioulis et al. \cite{3D60} adopt the extra coordinate feature in the equirectangular projection (ERP) domain for panoramas. PanoSUNCG\cite{PanoSUNCG} estimate omnidirectional depth and camera poses from 360$^\circ$ videos. Many approaches\cite{Wang2020bifuse, Jiang2021unifuse,Wang2023bifuse2,Feng2022segfuse} combine the ERP and CubeMap projection to overcome the distortion of panoramas. OmniFusion\cite{Li2022omnifusion} transforms the panorama into less-distorted perspective patches for depth estimation.

\textbf{Binocular omnidirectional depth estimation.} 360SD-Net\cite{360SD-Net} follows the stereo matching pipeline to estimate omnidirectional depth in the ERP domain for up-down stereo pairs. CSDNet\cite{Li2021csdnet} focuses on the left-right stereo and uses spherical CNNs to solve the distortions and proposes a cascade framework for accurate depth maps.

\textbf{Multi-view omnidirectional depth estimation.} Li et al. \cite{Li2022mode,Li2024robustmode} and Chiu et al. \cite{Chiu2023360MVSNet} use multiple panoramas as input to estimate 360$^\circ$ depth maps. Won et al. introcuce the spherical sweeping method and propose a series of algorithms\cite{SweepNet,Omnimvs,Omnimvs_J}, which build cost volume of multi-view fisheye images and estimate spherical depth via cost aggregation. Crown360 \cite{Komatsu2020crown360} uses icosahedron to represent the spherical information and leverages icosahedral CNN to estimate omnidrectional depth maps. Some methods \cite{Su2023omni} leverage cascade architectures for cost regularization to achieve high accuracy for omnidirectional detph estimation. Our previous work CasOmniMVS \cite{Wang2024Casomnimvs} proposes a multi-stage cascade network that dynamically adjusts the spherical sweeping range based on the predicted depth distribution from the previous stage to reduce the mismatches and improve the depth accuracy. S-OmniMVS\cite{chenSOmniMVS2023} focuses on the spherical geometry to deal with the fisheye distortion of input images and the omnidirectional distortion in cost aggregation. OmniVidar\cite{OmniVidar} adopts the triple sphere camera model and rectifies the multiple fisheye images into stereo pairs of four directions to obtain depth maps.

There are also some methods focus on the optimization of inference speed. Meuleman et al. \cite{realtimeOmni} propose an adaptive spherical matching and an efficient cost aggregation method to achieve real-time omnidirectional MVS. FastOmniMVS\cite{Wang2023FastOmniMVS} also adopts a lightweight architecture and leverage quantization aware training for accelaration on edge devices. Romnistereo\cite{Jiang2024Romnistereo} proposes a recurrent omnidirectional stereo matching algorithm to optimize 360$^\circ$ depth maps iteratively.

Most existing algorithms leverage the synthetic datasets proposed by Won et al. \cite{SweepNet,Omnimvs,Omnimvs_J} for training, some methods focus on the training strategies on real-world data. Lee et al. \cite{leeSemiSupervised2022} uses selective loss that combines photometric re-projection loss of images and the supervision with sparse LiDAR poinclouds. Chen et al. \cite{Chen2023UnOmnimvs} reconstruct two panoramas from different sets of input images based on the predicted depth and use pseudo-stereo loss for model training.

In summary, existing methods primarily focus on accuracy performance on synthetic datasets, with only a few studies exploring real-time algorithms on edge computing platforms for real-world applications. Moreover, research on training strategies using unlabeled real-world data remains insufficient, posing challenges for generalization in complex environments.

\subsection{Deep Learning based Stereo Matching}
MCCNN\cite{zbontar2016mccnn} first implements the feature extraction with CNNs. Many methods\cite{PSMNet,Shen2021CFNet,Xu2022acvnet,Zeng2024DMCA-Net} construct 3D cost volume with image features and optimize the 3D-CNN based cost aggeration to improve the accuracy. Some approaches\cite{Mayer2016dispnet,Xu2020aanet,Song2020edgestereo} compute the 2D feature correlation volume for efficient performance. Recently, some methods\cite{lipson2021raft, Li2022crestereo} leverage recurrent unit to estimate disparity iteratively. CREStereo\cite{Li2022crestereo} designs a hierarchical network to update disparities iteratively and proposes an adaptive group correlation layer to achieve state-of-the-art (SOTA) performance.

\section{Method}

\begin{figure}[]%
  \centering
  \includegraphics[width=0.7\textwidth]{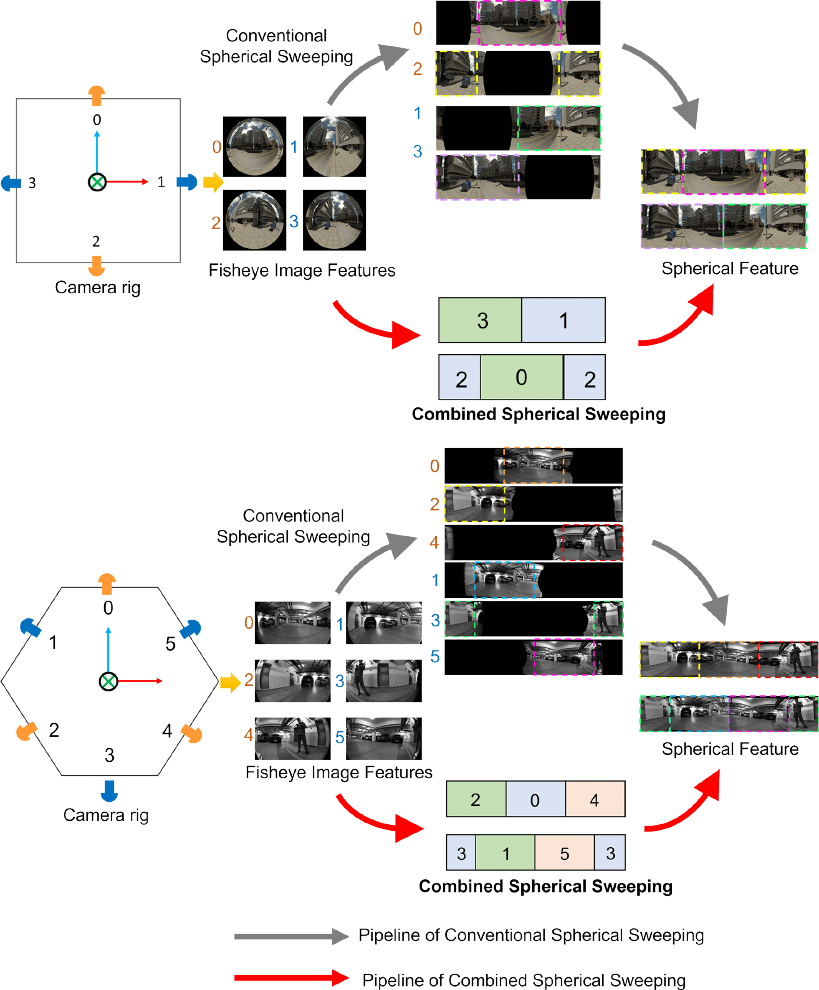}
  \caption{The proposed Combined Spherical Sweeping and the comparison with conventional method. Conventional method is indicated by the gray arrows, which projects the featrure map of every input image onto the completed sphere and then stitches the features to construct two spherical features. The proposed method is illustrated by the red arrows, directly projects multi-view features into two spherical features, significantly reducing the computational cost of the projection process}\label{fig:combined_sweeping}
\end{figure}

\subsection{Real-time Omnidirectional Depth Estimation}
Existing methods achieve omnidirectional depth estimation from multi-view fisheye image inputs through the pipeline that includes feature extraction from input images, construction of omnidirectional matching cost volume, cost aggregation and depth regression. Generally, these methods follows the spherical sweeping algorithms proposed by OmniMVS\cite{Omnimvs} to build the omnidirectional cost volume, which projects features from multi-view fisheye images onto a set of hypothetical spheres at different depths. Many algorithms apply 3D-CNNs for cost volume aggregation to improve the accuracy. However, the extensive projection and interpolation operations and the usage of 3D-CNNs limit the computational efficiency of the algorithm on edge computing platforms. Therefore, this paper proposes an improved Combined Spherical Sweeping method and an optimized network architecture to achieve real-time omnidirectional depth estimation on edge computing platforms.

\subsubsection{Combined Spherical Sweeping Method}

\begin{figure}[]%
  \centering
  \includegraphics[width=0.8\textwidth]{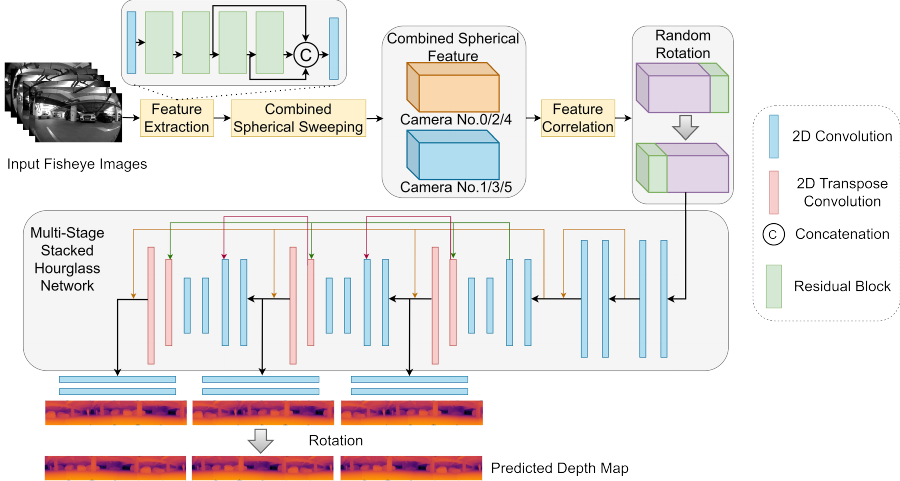}
  \caption{The model structure of proposed Rt-OmniMVS. The proposed method utilizes Combined Spherical Sweeping to construct omnidirectional matching costs based on features of multi-view fisheye images, followed by cost aggregation to predict depth. The random rotation is leveraged to improve the performance. The model employs a lightweight structural design and multi-scale supervision}\label{fig:model_structure}
\end{figure}

Most of MODE algorithms follow the spherical sweeping method proposed by OmniMVS\cite{SweepNet,Omnimvs,Omnimvs_J} to build the matching cost of objects at different depths via image features projection. The feature projection involves numerous complex matrix indexing and interpolation operations, which increases runtime on edge computing systems, becoming a bottleneck that hinders real-time algorithm performance. Due to the limited FoV of the camera, each 360$^\circ$ spherical feature map contains some invalid regions. Some methods \cite{Omnimvs_J,Jiang2024Romnistereo} have optimized spherical sweeping by improving the feature fusion strategies for areas outside the FoV of each camera. As indicated by the gray arrows in Fig .\ref{fig:combined_sweeping}, the features of each input image are individually projected onto 360$^\circ$ spheres, and then stitched to complete spherical features for calculation of matching cost at different depths. The improvements in existing methods reduce the negative impact of invalid regions on matching costs but do not effectively decrease the computational burden of projection interpolation in the spherical sweeping process.

Although edge computing platforms have been optimized for common operations such as 2D convolution, significant challenges still remain in computational efficiency for interpolation operations involving extensive memory copying (e.g., grid sample) and computationally intensive modules such as 3D convolutions. During the construction of cost volume with spherical sweeping, numerous operations such as random memory access and memory copy are required, which has a significant impact on the inference speed of edge platforms. These limitations have become bottlenecks restricting the real-time deployment of spherical-sweeping-based methods on edge devices.

Therefore, we introduce a Combined Spherical Sweeping method that can significantly reduce the number of indexing and interpolation operations in matching cost building, thereby accelarate the inference speed. As indicated by the red arrows in Fig. \ref{fig:combined_sweeping}, based on the layout and FoV of the cameras, all input images can be divided into two distinct groups to form two complete 360$^\circ$ spherical features. The orientation and FoV of cameras in each group can cover the 360$^\circ$ region. We reconstruct the projection mapping table with the parameters of hypothetical spheres and camera system. The feature of each camera is then directly projected into the combined spherical feature map based on the reconstructed mapping table, requiring only the projection of two spherical features.

Given $N$ camera inputs and $D$ hypothetical spheres, the conventional spherical sweeping projects every input feature map onto each sphere, which requires $N \times D$ projection operation computations. The proposed Combined Spherical Sweeping method only requires to build two spherical features for each hypothetical sphere, which can reduce the amount the computations to $2 \times D$. Therefore, the number of projection operations of propsoed combined spherical sweeping is reduced to $\frac{2}{N}$ of the original amount. As the samples shown in Fig. \ref{fig:combined_sweeping}, under configurations of four and six fisheye cameras respectively, the proposed Combined Spherical Sweeping method effectively reduces the amount of feature projection computations to $\frac{1}{2}$ and $\frac{1}{3}$ of the original approach, correspondingly.

The proposed Combined Spherical Sweeping computes projection interpolation tables based on the layout and FoV of the cameras. This enables the integration of spherical features during the projection process, significantly reducing the computational cost of the interpolation process, making it well-suited for deployment on edge computing platforms. Furthermore, the proposed method can adapt to various camera configurations based on intrinsic and extrinsic parameters, ensuring compatibility with diverse sensor layouts, while the percentage of computational reduction is relative to the camera configuration. The proposed method decouples the camera number from the model architecture, allowing the construction of two complete spherical feature representations and matching cost volumes based on the Combined Spherical Sweeping with any number of input cameras, thereby facilitating transfer to applications with different camera configurations.

\subsubsection{Lightweight Network Architecture}
In this paper, we propose Rt-OmniMVS, a real-time omnidirectional depth estimation algorithm for real-scene applications on edge computing devices. The model architecture is illustrated in Fig. \ref{fig:model_structure}. The model first extracts features of input fisheye images. Then we employ Combined Spherical Sweeping and project multi-view images onto hypothetical spheres at different depths within a central coordinate system. We build two sets of complete 360$^\circ$ features at different hypothetical spheres. The matching cost is calculated via cosine similarity of two sets of spherical features, inspired by correlation calculation methods in lightweight stereo matching methods\cite{Mayer2016dispnet,Xu2020aanet}. The dimensions of the matching cost is $D\times H\times W$, where $D$ represents the number of hypothetical spherical surfaces, and $H$ and $W$ represent the height and width of the feature map, respectively. We implement a multi-stage hourglass network with 2D CNNs to aggregate and regularize the matching cost and predict multi-stage 360$^\circ$ depth maps. We use multi-stage $smoothL_1$\cite{Girshick2015smoothl1} loss function:
\begin{align}
  Loss = \sum_{i=1}^{3}\lambda_i*smoothL_1(\hat{y}, y_i)
  \label{eq_loss}
\end{align}
where $\hat{y}$ denotes the groundtruth depth and $y_{i|i=1,2,3}$ denote the predicted depth of three stages. To avoid overfitting to the camera layout and orientation, we apply random horizontal rotations to the matching cost for aggregation and then rotate the predicted depth map back to its original orientation.

\subsection{Teacher-student Learning with Unlabeled Data}
Most omnidirectional depth estimation methods are typically trained and validated with synthetic datasets introduced by OmniMVS \cite{Omnimvs}, which suffer from limited scenes diversity and lack of real-world data. Incorporating real-world scene data into training can significantly improve the accuracy and generalization performance of the algorithm. Considering the challenges of obtaining accurate depth groundtruth in real-world, we propose a teacher-student learning framework that uses the teacher model generates pseudo-labels as groundtruth depth maps to train the student model effectively combined with data and model augmentation techniques.

OmniVidar\cite{OmniVidar} predicts local depth map via stereo matching in different directions and stitches the predicted depth into 360$^\circ$ depth map. Inspired by this, we leverage the recent SOTA stereo matching method CREStereo\cite{Li2022crestereo}, which is reported achieving high accuracy in real-world scenarios, as the teacher model to generate pseudo labels for the training of proposed Rt-OmniMVS.

\begin{figure}[]%
  \centering
  \includegraphics[width=1.0\textwidth]{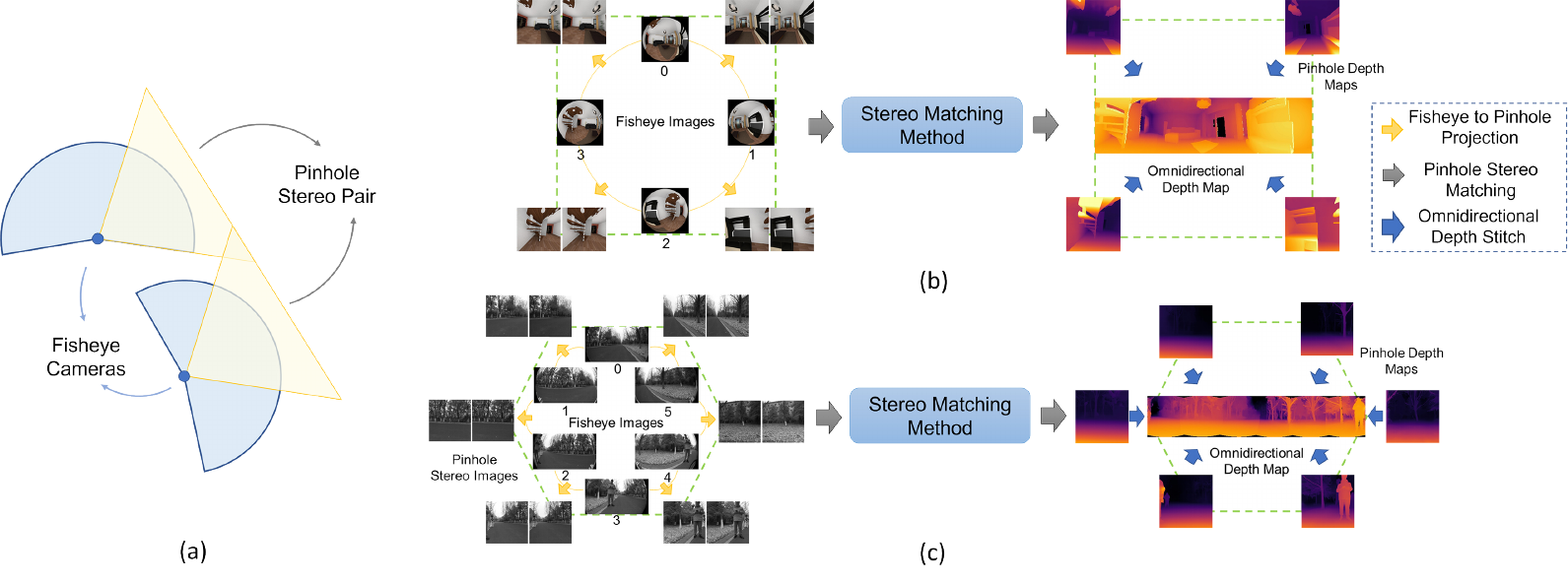}
  \caption{The diagram of proposed pseudo depth generation method. Multi-view fisheye images are projected into pinhole stereo image pairs in various directions to obtain depth maps based on stereo matching, and stitched together to construct a omnidirectional depth map. (a) presents the process of image projection and the generation of pinhole stereo pairs. (b) and (c) demonstrate the generation process of pseudo labels for camera systems with four and six fisheye cameras as input, respectively}\label{fig:pseudo_depth}
\end{figure}

Fig. \ref{fig:pseudo_depth} illustrates the process to predict omnidirectional depth pseudo-labels based on the stereo matching model. As shown in Fig. \ref{fig:pseudo_depth}(a), we generate a pinhole stereo image pair for every two adjacent fisheye cameras via image projection, based on the intrinsic and extrinsic parameters. The virtual pinhole cameras use idealized camera model with both horizontal and vertical FoV set at 75$^\circ$. For a surround-view imaging system comprised $N$ fisheye cameras, each camera forms stereo pairs with two adjacent cameras, resulting in $N$ binocular image pairs with varying orientations. Collectively, these pairs ensure comprehensive 360$^\circ$ coverage. We then use a pretrained stereo matching algorithm to predict the depth map for each pinhole stereo pair at different orientations. Finally, we stitch and fuse these $N$ depth maps to obtain a high-precision 360$^\circ$ depth map. Fig. \ref{fig:pseudo_depth}(b) and (c) illustrate the pseudo-label generation process under two different camera configurations: the four-fisheye camera layout from the dataset proposed by OmniMVS \cite{Omnimvs} and the six-fisheye camera setup proposed in this study.

\begin{figure}[]%
  \centering
  \includegraphics[width=0.5\textwidth]{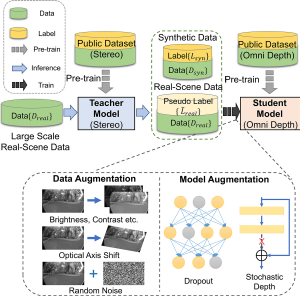}
  \caption{The diagram of proposed teacher-student learning strategy. The student model is first pre-trained on the public synthetic dataset, and then trained with pseudo-labels inferred by the teacher model, while applying data and model augmentation to enhance performance}\label{fig:self_training}
\end{figure}

Fig. \ref{fig:self_training} present the proposed teacher-student self-training strategy. The teacher model CREStereo\cite{Li2022crestereo} is trained on public stereo datasets to achieve high accuracy, and then used to inference omnidirectional depth pseudo-labels for real-world scenarios data, as shown in Fig. \ref{fig:pseudo_depth}. We also generate a synthetic dataset and mix it with the real-scene data to build the hybrid dataset. The student model Rt-OmniMVS is initially trained on the public OmniThings dataset \cite{Omnimvs}, and then trained on the hybrid dataset to enhance the accuracy and robustness for real applications.

To improve the accuracy and generalization of the lightweight model, we employ \textbf{data augmentation (DA)} and \textbf{model augmentation (MA)} strategies during training. As shown in Fig. \ref{fig:self_training}, for data augmentation, in addition to common techniques such as brightness and contrast adjustments, we develop \textbf{random noise} and \textbf{optical axis shift} augmentations. The random noise method adds random Gaussian or Poisson noise to input images. The optical axis shift augmentation applies small-scale random affine transformations to the images, introducing geometric errors to overcome the misalignments in camera calibration. For model augmentation, we utilize techniques such as Dropout that randomly deactivates neurons and Stochastic Network Depth that randomly deactivates the forward path of residual blocks during training. The MA methods reduce overfitting and enhance the generalization of the model. In summary, the proposed self-training method leverages pseudo-labels generated by the teacher model to guide the training of a student model. The DA and MA methods effectively improve the capacity of the student model to achieve high accuracy and robustness on real scenes.

\subsection{Real-time Omnidirectional Depth Estimation System}

\begin{figure}[]%
  \centering
  \includegraphics[width=0.5\textwidth]{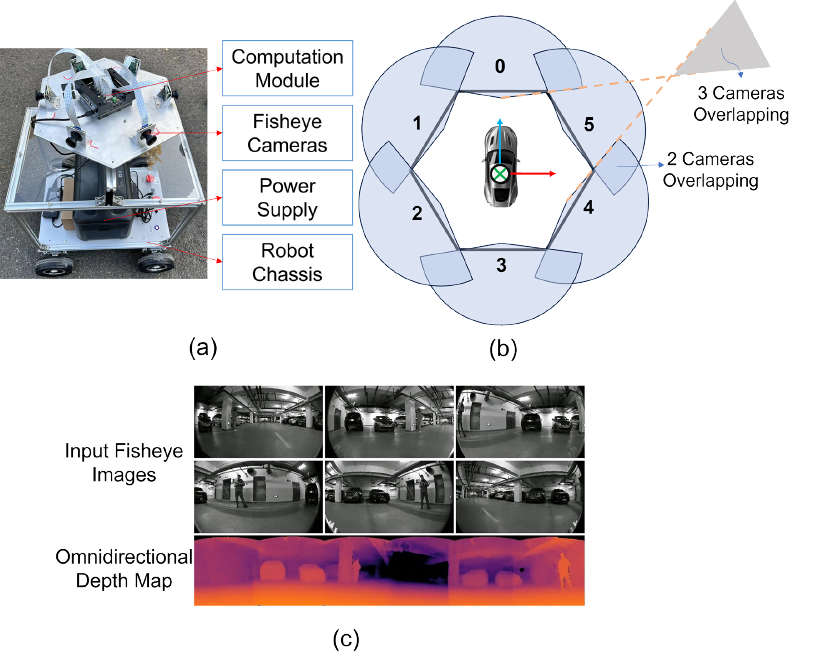}
  \caption{Overview of the proposed multi-view omnidirectional depth estimation system HexaMODE. (a) The hardware structure and the prototype of proposed system. (b) The rig of six fisheye cameras and the indication of multi-view overlapping regions. (c) Sample of input images and predicted depth in real scene}\label{fig:system_overview}
\end{figure}

To validate the deployment and real-time inference of the proposed Rt-OmniMVS on edge computing platforms, as well as the accuracy and generalization performance in complex real-world scenarios, this study constructs a multi-view omnidirectional depth estimation system tailored for real-world applications. As depicted in Fig. \ref{fig:system_overview}(a), the proposed HexaMODE system is constructed on a robot chassis, consisting of a computational module, a fisheye camera system and a power supply module. We employ a surround-view system composed of six fisheye cameras, which expands the overlapping area of different cameras while balancing the computational complexity of the algorithm. The overall system has dimensions of approximately 0.6m (meters) in length, 0.55m in width, and 0.73m in height. We use one NVIDIA Jetson AGX Orin to control the system and run the omnidirectional depth estimation algorithm. Fig. \ref{fig:system_overview}(b) shows the layout of cameras. In the design, the six fisheye cameras are arranged in a regular hexagonal pattern, with an azimuthal orientation difference of 60$^\circ$ between adjacent cameras, and an optical center distance of 0.17m. Due to intrinsic parameter variations and potential installation errors, the extrinsic parameters of each camera are obtained via calibration. Each fisheye camera has a horizontal FoV (Field of View) of 161$^\circ$ and a vertical FoV of 75$^\circ$. The configuration of cameras ensure 360$^\circ$ surrounding coverage and provide overlapping regions between camera views for feature matching, enabling accurate depth estimation. As illustrated in the Fig. \ref{fig:system_overview}, comparing with the four-camera system in previous works \cite{Omnimvs}, the proposed HexaMODE system with six fisheye cameras enables the coverage of regions by the FoV of three cameras, leading to the establishment of more confidencial geometric constraints, especially for the occlusion areas and near-fields objects.

\subsection{Hexa360Depth Dataset}
\begin{figure}[]%
  \centering
  \includegraphics[width=0.8\textwidth]{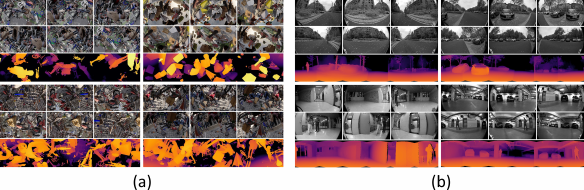}
  \caption{The synthetic data (a) and real-scene data (b) samples of the proposed Hexa360Depth Dataset. Each sample presents six input images and the groundtruth depth map(synthetic) or pseudo groundtruth(real)}\label{fig:dataset}
\end{figure}

We propose a hybrid dataset consisting of synthetic data with groundtruth labels and real-world data with pseudo labels to train the proposed model, enabling high-accuracy and high-generalization performance in real-world scenarios. The synthetic dataset is generated with the Carla simulator, following the camera layout of proposed HexaMODE system. We employ various backgrounds and random objects of different sizes and positions for diverse data. The real-world dataset is collected by the proposed HexaMODE system in different environments such as indoor, outdoor, roadways, and parking lots, etc. We follow the pipeline shown in Fig. \ref{fig:pseudo_depth} to generate pseudo labels for real-world data. By generating pseudo-labels for unlabeled real-world data for model training and validation, the cost of real-world data acquisition can be significantly reduced. This approach enables the construction of large-scale datasets, thereby enhancing the generalization performance of the algorithm. Fig. \ref{fig:dataset} presents the input images and depth ground truth from the synthetic random object dataset, as well as the input images and pseudo-groundtruth labels from various categories of real-world scenes.

The total dataset comprises 33 scenes and 41281 samples, as summarized in Table \ref{tab:dataset_collect}. The synthetic random obejects dataset contains 9 different scenes and 14003 samples of data. The real-world scene data consist of four categories of datasets, comprising a total of 24 distinct scenes and 27278 data samples. We divide the hybrid dataset into a training set and a test set. The training set consists of 35140 samples, including 23137 real-world samples and 12003 synthetic smaples. While the test set comprises 2000 samples from the simulated data and 4141 samples from the real-world data (a total of 6141). The data in the training and testing sets are sourced from different scenes, exhibiting differences in aspects such as scene content and depth distribution. The proposed Hexa360Depth dataset comprises diverse random objects and real-world scenes with varying environmental conditions and depth distributions. It effectively supports the training of models in feature extraction and multi-view matching, offering practical potential for multi-view omnidirectional depth estimation.

\begin{table}[]
  \centering
  \caption{Summary of the proposed Hexa360Depth Dataset}
  \label{tab:dataset_collect}
  \resizebox{0.5\textwidth}{!}{%
    \begin{tabular}{cc|cc}
      \toprule
      Data Type                             & Category            & Num of Scenes  & Num of Samples \\ \toprule
      Synthetic                             & Random Objects      & 9              & 14003          \\ \midrule
      \multirow{5}{*}{Real Scene}           & Outdoor Road        & 8              & 10753          \\
                                            & Outdoor Parking     & 6              & 6868           \\
                                            & Indoor Hallway      & 4              & 3804           \\
                                            & Underground Parking & 6              & 5853           \\ \cline{2-4}
                                            & Summary(Real data)  & 24             & 27278          \\ \toprule
      \multicolumn{2}{c|}{\textbf{Summary}} & \textbf{33}         & \textbf{41281}                  \\ \bottomrule
    \end{tabular}
  }
\end{table}

\section{Experiment}

\subsection{Implementation Details}
We implement and train the proposed Rt-OmniMVS with Pytorch framework. We export the model to ONNX model file and then utilize the NVIDIA TensorRT toolchain to deploy the model on the NVIDIA Orin platform. The model is first pre-trained for 40 epochs on the OmniThings dataset with a initial learning rate of 0.0005, and decays the learning rate to 80$\%$ of former value for every 10 epochs. Subsequently, the model is trained for 10 epochs on the proposed hybrid dataset with a initial learning rate of 0.001, and decays the learning rate to 50$\%$ of former value for every 2 epochs. The coefficients $\lambda_{i|i=1,2,3}$ of the loss in Eq.\ref{eq_loss} are set to 0.5, 0.7, and 1.0, respectively. We set the number of hypothetical spheres to 64, and set depth range to $[1.65m,1000m]$ for OmniThings and $[0.5m,100m]$ for proposed hybrid dataset.

\subsection{Evaluation Metrics}
We use commonly metrics in depth estimation to evaluate the algorithm, including MAE(mean absolute error), RMSE(root mean square error), AbsRel(absolute relative error), SqRel(square relative error), SILog(scale-invariant logarithmic error) \cite{Silog}, $\delta 1,2,3$(accuracy with threshold that $\max(\frac{\hat{y}}{y},\frac{y}{\hat{y}}) < 1.25,1.25^2,1.25^3$) \cite{Delta123}. Higher values are better for the accuracies $\delta 1,2,3$, while lower values are better for other error metrics.

\subsection{Evaluation on Public Datasets}

We first evaluate the proposed Rt-OmniMVS on public datasets OmniThings, OmniHouse and Urban(Sunny, Cloudy and Sunset) \cite{Omnimvs} with four input fisheye cameras. Rt-OmniMVS has a different setting of hypothetical spheres with other methods. Therefore, we convert the results into absolute depth values for evaluation. We compare the proposed algorithm with recent methods using depth error metrics. All the methods are trained on OmniThings dataset and then finetuned on OmniHouse and Sunny. Table \ref{tab:rtomnimvs_omnithings_fullres} and \ref{tab:rtomnimvs_omnihouse_urban_fullres} presents the evaluation results and comparison, along with the GPU memory usage and inference time of each method. We run the experiments with one NVIDIA 5090D GPU. The results indicate that our method achieves second-best performance across many metrics. The proposed Rt-OmniMVS achieves second-best result on Omnithings datasets, and also surpasses Romnistereo \cite{Jiang2024Romnistereo} and OmniStereo \cite{Deng2025OmniStereo} in some metrics (i.e. AbsRel, SqRel, etc.) on OmniHouse and Urban datasets.With only minor losses in accuracy, the proposed Rt-OmniMVS significantly reduces inference time and memory consumption.

\begin{table}[]
  \centering
  \caption{Quantitative depth estimation results of proposed Rt-OmniMVS on OmniThings dataset. The metrics refer to depth errors}
  \label{tab:rtomnimvs_omnithings_fullres}
  \resizebox{\textwidth}{!}{%
    \begin{tabular}{cc|cccccccc|cc}
      \hline
      Datasets                    & Methods                                        & MAE$\downarrow$ & RMSE$\downarrow$ & AbsRel$\downarrow$ & SqRel$\downarrow$ & rSILog$\downarrow$ & $\delta1(\%)\uparrow$ & $\delta2(\%)\uparrow$ & $\delta3(\%)\uparrow$ & \makecell{Mem       \\(GB)} & \makecell{Time\\(ms)} \\ \hline
      \multirow{5}{*}{Omnithings} & OmniMVS \cite{Omnimvs}                         & 2.363           & 7.883            & 0.283              & 3.545             & 0.333              & 83.113                & 90.544                & 94.703                & 7.7           & 109 \\
                                  & Romnistereo \cite{Jiang2024Romnistereo}        & 2.221           & 6.318            & 0.231              & 2.850             & 0.304              & 81.680                & 89.561                & 94.585                & 3.9           & 45  \\
                                  & OmniStereo \cite{Deng2025OmniStereo}           & 2.745           & 8.162            & 0.488              & 8.317             & 0.368              & 82.296                & 90.558                & 94.214                & 5.4           & 31  \\
                                  & \makecell{CasOmniMVS \cite{Wang2024Casomnimvs}                                                                                                                                                                                                  \\(ours previous)} & 0.949           & 2.018            & 0.060              & 0.041             & 0.135              & 89.609                & 94.413                & 96.891                & 3.0           & 80  \\
                                  & \makecell{Rt-OmniMVS                                                                                                                                                                                                                            \\(ours)}  & 1.346           & 2.552            & 0.095              & 0.038             & 0.181              & 88.139                & 94.834                & 96.654                & 1.2           & 24  \\ \hline
    \end{tabular}
  }
\end{table}

\begin{table}[]
  \centering
  \caption{Quantitative depth estimation results of proposed Rt-OmniMVS on OmniHouse and Urban dataset. The metrics refer to depth errors}
  \label{tab:rtomnimvs_omnihouse_urban_fullres}
  \resizebox{0.85\textwidth}{!}{%
    \begin{tabular}{cc|cccccccc}
      \hline
      Datasets                        & Methods                                 & MAE$\downarrow$ & RMSE$\downarrow$ & AbsRel$\downarrow$ & SqRel$\downarrow$ & rSILog$\downarrow$ & $\delta1(\%)\uparrow$ & $\delta2(\%)\uparrow$ & $\delta3(\%)\uparrow$ \\ \hline
      \multirow{5}{*}{Omnihouse}      & OmniMVS \cite{Omnimvs}                  & 0.631           & 2.292            & 0.044              & 0.086             & 0.097              & 97.253                & 98.857                & 99.417                \\
                                      & Romnistereo \cite{Jiang2024Romnistereo} & 2.028           & 3.986            & 0.140              & 0.595             & 0.150              & 93.051                & 97.259                & 98.341                \\
                                      & OmniStereo \cite{Deng2025OmniStereo}    & 0.599           & 1.970            & 0.043              & 0.166             & 0.087              & 97.773                & 99.108                & 99.538                \\
                                      & CasOmniMVS \cite{Wang2024Casomnimvs}    & 0.497           & 1.321            & 0.029              & 0.013             & 0.063              & 97.758                & 98.972                & 99.387                \\
                                      & Rt-OmniMVS (ours)                       & 0.724           & 1.802            & 0.050              & 0.021             & 0.104              & 95.639                & 98.559                & 99.414                \\ \hline
      \multirow{5}{*}{Urban (Sunny)}  & OmniMVS \cite{Omnimvs}                  & 1.774           & 6.990            & 0.083              & 0.308             & 0.209              & 94.750                & 96.900                & 97.891                \\
                                      & Romnistereo \cite{Jiang2024Romnistereo} & 3.473           & 9.312            & 0.118              & 0.270             & 0.211              & 88.794                & 95.734                & 98.045                \\
                                      & OmniStereo \cite{Deng2025OmniStereo}    & 1.720           & 6.638            & 0.104              & 0.527             & 0.216              & 94.277                & 96.816                & 97.915                \\
                                      & CasOmniMVS \cite{Wang2024Casomnimvs}    & 1.471           & 5.736            & 0.051              & 0.088             & 0.145              & 92.416                & 95.394                & 96.741                \\
                                      & Rt-OmniMVS (ours)                       & 2.528           & 6.861            & 0.098              & 0.146             & 0.210              & 89.748                & 96.002                & 97.889                \\ \hline
      \multirow{5}{*}{Urban (Cloudy)} & OmniMVS \cite{Omnimvs}                  & 1.733           & 6.945            & 0.080              & 0.297             & 0.206              & 94.961                & 97.012                & 97.955                \\
                                      & Romnistereo \cite{Jiang2024Romnistereo} & 3.596           & 9.642            & 0.211              & 0.280             & 0.227              & 88.510                & 95.306                & 97.834                \\
                                      & OmniStereo \cite{Deng2025OmniStereo}    & 1.684           & 6.509            & 0.100              & 0.490             & 0.213              & 94.396                & 96.878                & 97.967                \\
                                      & CasOmniMVS \cite{Wang2024Casomnimvs}    & 1.439           & 5.419            & 0.051              & 0.083             & 0.143              & 92.552                & 95.526                & 96.741                \\
                                      & Rt-OmniMVS (ours)                       & 2.687           & 7.153            & 0.099              & 0.126             & 0.216              & 89.046                & 95.604                & 97.736                \\ \hline
      \multirow{5}{*}{Urban (Sunset)} & OmniMVS \cite{Omnimvs}                  & 1.773           & 6.978            & 0.085              & 0.314             & 0.209              & 94.691                & 96.881                & 97.892                \\
                                      & Romnistereo \cite{Jiang2024Romnistereo} & 3.427           & 9.213            & 0.121              & 0.273             & 0.214              & 89.105                & 95.863                & 98.065                \\
                                      & OmniStereo \cite{Deng2025OmniStereo}    & 1.698           & 6.637            & 0.104              & 0.524             & 0.219              & 94.341                & 96.823                & 97.886                \\
                                      & CasOmniMVS \cite{Wang2024Casomnimvs}    & 1.465           & 5.584            & 0.054              & 0.103             & 0.149              & 92.195                & 95.213                & 96.456                \\
                                      & Rt-OmniMVS (ours)                       & 2.597           & 6.950            & 0.102              & 0.154             & 0.215              & 89.402                & 95.826                & 97.835                \\ \hline
    \end{tabular}
  }
\end{table}

\subsection{Evaluation on Real-scene Datasets}
\begin{figure}[]%
  \centering
  \includegraphics[width=0.5\textwidth]{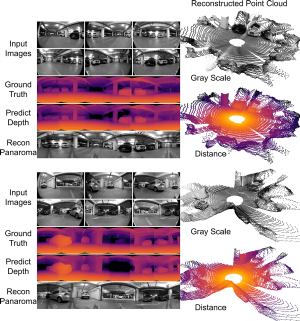}
  \caption{Qualitative results of HexaMODE on real-world indoor scenes}\label{fig:results_indoor}
\end{figure}

\begin{figure}[]%
  \centering
  \includegraphics[width=0.5\textwidth]{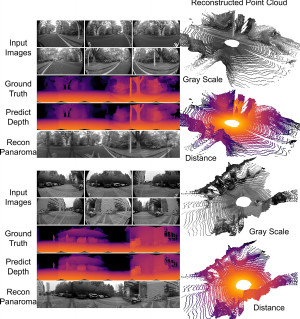}
  \caption{Qualitative results of HexaMODE on real-world outdoor scenes}\label{fig:results_outdoor}
\end{figure}

We evaluate the performance of proposed Rt-OmniMVS on the real-scene test set. Fig. \ref{fig:results_indoor} and Fig. \ref{fig:results_outdoor} show the qualitative results of the model on indoor and outdoor scenes, respectively. The figures display the input six fisheye images, the predicted depth maps, the depth pseudo-labels and the reconstructed panoramas at the central view via depth based image projection. Additionally, we reconstruct the 3D point cloud of the scene utilizing the predicted depth maps. Fig. \ref{fig:results_indoor} and Fig. \ref{fig:results_outdoor} display the point clouds rendered using grayscale values from the input images and pseudo color based on the distance from the system. The qualitative results of predicted depth maps and point clouds demonstrate that the proposed method achieves high-precision omnidirectional depth estimation and robust 3D structure reconstruction in real-world scenarios, with robustness across diverse indoor and outdoor environments.

We adjust the number of channels in the first layer of the cost aggregation module in OmniMVS \cite{Omnimvs} for six cameras input, and then train and test the model on the proposed Hexa360Depth dataset, with results presented in Table \ref{tab:rtomnimvs_compare_hexa}.
The experiment results of OmniMVS \cite{Omnimvs} demonstrate that propsoed dataset effectively supports model training. OmniMVS \cite{Omnimvs} performs better since the more complex network structure and parameters, but faces great challenges of deployment on edge devices. Our method offers greater advantages in deployment on edge computing platforms and real-time inference, while also achieves acceptable accuracy for low-speed applications.

We evaluate the effects of the proposed teacher-student training paradigm on the model performance in real-world scenarios. Table \ref{tab:rtomnimvs_res} presents a quantitative comparison of results under different training data and strategies settings. We select the model trained on the OmniThings dataset (Omni-pretrained) as the baseline for comparison. In Table \ref{tab:rtomnimvs_res}, "Syn" and "Real" represent the proposed synthetic data and real-world data, respectively. "DA" and "MA" denote the data augmentation and model augmentation training strategies. The test results are based on scenes within a 10-meter range. The comparison results indicate that as components are added to the training strategy, the accuracy generally shows a gradual improvement, confirming the effectiveness of the proposed training approach. The results reveal that after fine-tuning the model on synthetic data, accuracy metrics significantly improve. The inclusion of real-world data also leads to a significant reduction in error metrics, demonstrating the positive effects of real-world data on improving the generalization and accuracy of the model. Additionally, incorporating model augmentation significantly imporve the quantitative metrics, indicating that the use of techniques like Dropout and Stochastic Depth during training contributes to enhance the capacity and generalization of the model. The evaluation results demonstrate that proposed Rt-OmniMVS can efficiently achieve high-precision 360$^\circ$ depth maps and dense 3D point clouds, meeting the omnidirectional 3D perception requirements for robot navigation and low-speed autonomous driving scenarios. As illustrated in Table \ref{tab:rtomnimvs_res}, the model directly only trained on four-camera dataset OmniThings (with depth range of 1.65-1000m) reports favorable performance in metrics such as RMSE and SqRel on the six-camera Hexa360Depth dataset (with depth range of 0.5-100m), demonstrating generalization capability of the proposed Rt-OmniMVS cross different depth range and camera settings.

\begin{table*}[]
  \centering
  \caption{Quantitative depth estimation comparison of proposed Rt-OmniMVS on Hexa360Depth}
  \label{tab:rtomnimvs_compare_hexa}
  \resizebox{0.9\textwidth}{!}{%
    \begin{tabular}{c|cccccccc}
      \toprule
      Methods               & MAE$\downarrow$ & RMSE$\downarrow$ & AbsRel$\downarrow$ & SqRel$\downarrow$ & rSILog$\downarrow$ & $\delta1(\%)\uparrow$ & $\delta2(\%)\uparrow$ & $\delta3(\%)\uparrow$ \\ \toprule
      OmniMVS\cite{Omnimvs} & 0.326           & 1.050            & 0.058              & 0.047             & 0.120              & 95.552                & 98.443                & 99.195                \\
      Rt-OmniMVS (ours)     & 0.690           & 1.757            & 0.126              & 0.077             & 0.184              & 87.006                & 96.759                & 98.454                \\ \bottomrule
    \end{tabular}
  }
\end{table*}

\begin{table*}[]
  \centering
  \caption{Quantitative depth estimation results of proposed Rt-OmniMVS on real-scene dataset. \textbf{Syn} and \textbf{Real} denote synthetic and real-scene data, respectively. \textbf{DA} and \textbf{MA} denote data augmentation and model augmentation. Depth range is set to within 10m. The best results are marked in bold and the second best results are marked in underline. The metrics refer to depth errors}
  \label{tab:rtomnimvs_res}
  \resizebox{0.9\textwidth}{!}{%
    \begin{tabular}{c|cccccccc}
      \toprule
      Training Data and Stratages & MAE$\downarrow$   & RMSE$\downarrow$  & AbsRel$\downarrow$ & SqRel$\downarrow$ & rSILog$\downarrow$ & $\delta1(\%)\uparrow$ & $\delta2(\%)\uparrow$ & $\delta3(\%)\uparrow$ \\ \toprule
      Omni-pretrained             & 0.973             & \textbf{1.437}    & 0.237              & 0.125             & 0.293              & 60.034                & 83.718                & 95.351                \\
      Omni+Syn                    & 0.864             & 2.202             & 0.179              & 0.249             & 0.368              & 83.551                & 93.694                & 96.377                \\
      Omni+Syn+Real               & 0.791             & 1.854             & 0.145              & \underline{0.087} & \underline{0.205}  & \underline{83.951}    & 94.851                & \underline{97.970}    \\
      Omni+Syn+Real+DA            & \underline{0.785} & 1.865             & \underline{0.144}  & 0.089             & 0.206              & 83.460                & \underline{95.115}    & 97.926                \\
      Omni+Syn+Real+DA+MA         & \textbf{0.690}    & \underline{1.757} & \textbf{0.126}     & \textbf{0.077}    & \textbf{0.184}     & \textbf{87.006}       & \textbf{96.759}       & \textbf{98.454}       \\ \bottomrule
    \end{tabular}
  }
\end{table*}

To comprehensively evaluate the performance in real-world scenes, we have supplemented the metrics under different depth thresholds (5m, 10m, 20m and 100m), as presented in Table \ref{tab:rtomnimvs_res_range}. The results indicate that accuracy decreases with increasing depth range. However, relative error AbsRel remains below 20$\%$ within a 100-meter range, demonstrating potential for low-speed practical applications.

\begin{table*}[]
  \centering
  \caption{Quantitative depth estimation results of proposed Rt-OmniMVS at different depth range settings}
  \label{tab:rtomnimvs_res_range}
  \resizebox{0.9\textwidth}{!}{%
    \begin{tabular}{c|cccccccc}
      \toprule
      Depth Threshold (meter) & MAE$\downarrow$ & RMSE$\downarrow$ & AbsRel$\downarrow$ & SqRel$\downarrow$ & rSILog$\downarrow$ & $\delta1(\%)\uparrow$ & $\delta2(\%)\uparrow$ & $\delta3(\%)\uparrow$ \\ \toprule
      5                       & 0.256           & 0.567            & 0.082              & 0.034             & 0.114              & 95.125                & 99.205                & 99.655                \\
      10                      & 0.690           & 1.757            & 0.126              & 0.077             & 0.184              & 87.006                & 96.759                & 98.454                \\
      20                      & 1.444           & 3.394            & 0.168              & 0.116             & 0.248              & 79.591                & 91.782                & 96.219                \\
      100                     & 1.961           & 4.312            & 0.178              & 0.119             & 0.272              & 77.199                & 89.974                & 95.059                \\ \bottomrule
    \end{tabular}
  }
\end{table*}

\subsection{Evaluation of Efficiency}

\begin{table*}[]
  \centering
  \caption{Inference time and metrics comparison of Rt-OmniMVS with different spherical sweeping methods}
  \renewcommand\arraystretch{1.2}
  \label{tab:rtomnimvs_infer_time}
  \resizebox{\textwidth}{!}{%
    \begin{tabular}{c|c|cccccccc}
      \toprule
      {Spherical Sweeping Method} & {\makecell{Time(ms)                                                                    \\ AGX Orin}} & MAE$\downarrow$ & RMSE$\downarrow$ & AbsRel$\downarrow$ & SqRel$\downarrow$ & rSILog$\downarrow$ & $\delta1(\%)\uparrow$ & $\delta2(\%)\uparrow$ & $\delta3(\%)\uparrow$\\ \toprule
      {Original \cite{Omnimvs_J}} & 201                 & 0.428 & 1.285 & 0.088 & 0.112 & 0.140 & 95.018 & 97.994 & 98.843 \\
      Combined (Ours)             & 65                  & 0.690 & 1.757 & 0.126 & 0.077 & 0.184 & 87.006 & 96.759 & 98.454 \\ \bottomrule
    \end{tabular}
  }
\end{table*}

\begin{table}[]
  \centering
  \caption{Computational complexity comparison of student model Rt-OmniMVS and teacher model CREStereo}
  \label{tab:teacher_student_compare}
  \resizebox{0.5\textwidth}{!}{%
    \begin{tabular}{c|c|c|c}
      \toprule
      Method & Param(MB) & TFLOPs & Time(ms/frame)  \\ \toprule
      \multirow{2}{*}{\makecell{Teancher Model      \\ CREStereo\cite{Li2022crestereo}}}      & \multirow{2}{*}{5.4} & \multirow{2}{*}{3.630} & {\makecell{513 \\ (Depth Only)}}           \\ \cline{4-4}
             &           &        & {\makecell{2704 \\(w/ Projection)}} \\
      \midrule
      \multirow{2}{*}{\makecell{Student Model       \\ Rt-OmniMVS}} & \multirow{2}{*}{5.2} & \multirow{2}{*}{0.496} & \multirow{2}{*}{65}       \\
             &           &        &                 \\
      \bottomrule
    \end{tabular}
  }
\end{table}

We compare the inference time and error methics of Rt-OmniMVS using different spherical sweeping methods on the NVIDIA Jetson AGX Orin platform, as shown in Table \ref{tab:rtomnimvs_infer_time}. It costs 201 ms/frame using the original method \cite{Omnimvs_J}. In contrast, the proposed Combined Spherical Sweeping reduces the inference time to 65 ms/frame, achieving real-time performance of more than 15 fps on edge computation devices. As illustrated in Table \ref{tab:rtomnimvs_infer_time}, the Combined Spherical Sweeping can reduce the inference time to less than 1/3 of original method, while the accuracy can still meet the demands of some low-speed applications such as robots or UAVs. The input image resolution is $960 \times 540$, and the output cropped depth map resolution is $960 \times 192$.

We also compare the efficiency of teacher model CREStereo\cite{Li2022crestereo} and student model Rt-OmniMVS. As detailed in Table \ref{tab:teacher_student_compare}, the teacher model and student model have similar amount of parameters. However, CREStereo\cite{Li2022crestereo} employs an iterative optimization method and divides the omnidirectional depth into six pairs of stereo matches, leading to higher computational demands and slower inference time. If the projection process shown in Fig. \ref{fig:pseudo_depth} is included, generating the depth for each frame takes more than 2.7 seconds. In contrast, the proposed Rt-OmniMVS directly predict the 360$^\circ$ scene depth with an optimized design, resulting in lower complexity and faster inference speed. Therefore, the training strategy employed in this work significantly reduces computational complexity and inference time while maintaining high algorithm accuracy.

\section{Conclusion}
In this paper, we introduce Combined Spherical Sweeping method and biuld a lightweight omnidirectional depth estimation method named Rt-OmniMVS to achieve real-time inference on the edge computing platform NVIDIA Orin. To achieve the high accuracy and high generalization of complex real world scenes, we introduce a teacher-student training strategy that leverage SOTA stereo matching method as teacher model to generate pseudo-labels for unlabeled real data to train the student model. Data augmentation and model augmentation methods are leveraged during training to enhance the generalization capability of Rt-OmniMVS model. To validate the performance on real-world scenarios and collect large-scale unlabeled real-world data, we build an omnidirectional depth estimation system HexaMODE. We collect a real-scene dataset on various scenarios using the system and generate pseudo depth groundtruth with SOTA stereo matching algorithm. Combined with the generated synthetic data, a hybrid dataset is constructed for model training and validation. Extensive experiments validate the high-accuracy and real-time performance of the proposed Rt-OmniMVS algorithm and HexaMODE system on real-world scenarios. The proposed Rt-OmniMVS not only surpasses existing approaches in inference speed on GPU platforms, but also achieves real-time performance on edge computing platforms. Additionally, experiments on different datasets demonstrate the flexibility and compatibility for various camera layouts. The study presents the potential applications of omnidirectional depth estimation in the fields of autonomous driving and robotics.

Although the proposed method incorporates relevant optimizations, some experiment results still exhibit insufficient horizontal continuity in the predicted omnidirectional depth. In future work, we will further refine the combined spherical sweeping method to enhance the ability to fuse features across the boundaries of different camera views, and also leverage the semantic consistency in multi-view overlapping regions, aiming to improve both the accuracy and cross view continuity of 360$^\circ$ depth prediction. We will further investigate novel and efficient network architecture designs to enhance algorithmic performance while satisfying real-time performance requirements.

\section*{Acknowledgement}
This work is supported by Suzhou Science and Technology Plan (Frontier Technology Research Project) SYG202334, the Startup Foundation for Introducing Talent of NUIST (2025r030), and the Sci-Tech Innovation Talent Program Project of the National Administration for Market Regulation (KJLJ202319).

\printcredits

\bibliographystyle{elsarticle-num}

\bibliography{main}

@inproceedings{PSMNet,
  author    = {Chang, J. and Chen, Y.},
  title     = {Pyramid Stereo Matching Network},
  booktitle = {2018 IEEE/CVF Conference on Computer Vision and Pattern Recognition, {CVPR}},
  pages     = {5410-5418},
  isbn      = {2575-7075},
  doi       = {10.1109/CVPR.2018.00567},
  year      = {2018}
}

@inproceedings{Mayer2016dispnet,
  author    = {Mayer, Nikolaus and Ilg, Eddy and Häusser, Philip and Fischer, Philipp and Cremers, Daniel and Dosovitskiy, Alexey and Brox, Thomas},
  booktitle = {2016 IEEE Conference on Computer Vision and Pattern Recognition, {CVPR}},
  title     = {A Large Dataset to Train Convolutional Networks for Disparity, Optical Flow, and Scene Flow Estimation},
  year      = {2016},
  volume    = {},
  number    = {},
  pages     = {4040-4048},
  doi       = {10.1109/CVPR.2016.438}
}

@inproceedings{Xu2020aanet,
  author    = {Xu, Haofei and Zhang, Juyong},
  booktitle = {2020 IEEE/CVF Conference on Computer Vision and Pattern Recognition, {CVPR}},
  title     = {AANet: Adaptive Aggregation Network for Efficient Stereo Matching},
  year      = {2020},
  volume    = {},
  number    = {},
  pages     = {1956-1965},
  doi       = {10.1109/CVPR42600.2020.00203}
}

@article{Song2020edgestereo,
  author    = {Xiao Song and
               Xu Zhao and
               Liangji Fang and
               Hanwen Hu and
               Yizhou Yu},
  title     = {EdgeStereo: An Effective Multi-task Learning Network for Stereo Matching
               and Edge Detection},
  journal   = {International Journal of Computer Vision},
  volume    = {128},
  number    = {4},
  pages     = {910--930},
  year      = {2020},
  doi       = {10.1007/s11263-019-01287-w}
}

@inproceedings{Xu2022acvnet,
  author    = {Gangwei Xu and
               Junda Cheng and
               Peng Guo and
               Xin Yang},
  title     = {Attention Concatenation Volume for Accurate and Efficient Stereo Matching},
  booktitle = {{IEEE/CVF} Conference on Computer Vision and Pattern Recognition, {CVPR}},
  pages     = {12971--12980},
  publisher = {{IEEE}},
  year      = {2022},
  doi       = {10.1109/CVPR52688.2022.01264}
}

@article{Li2021csdnet,
  title    = {Omnidirectional stereo depth estimation based on spherical deep network},
  journal  = {Image and Vision Computing},
  volume   = {114},
  pages    = {104264},
  year     = {2021},
  issn     = {0262-8856},
  doi      = {https://doi.org/10.1016/j.imavis.2021.104264},
  author   = {Ming Li and Xuejiao Hu and Jingzhao Dai and Yang Li and Sidan Du}
}

@inproceedings{Wang2020bifuse,
  author    = {Wang, Fu-En and Yeh, Yu-Hsuan and Sun, Min and Chiu, Wei-Chen and Tsai, Yi-Hsuan},
  booktitle = {2020 IEEE/CVF Conference on Computer Vision and Pattern Recognition, {CVPR}},
  title     = {BiFuse: Monocular 360 Depth Estimation via Bi-Projection Fusion},
  year      = {2020},
  volume    = {},
  number    = {},
  pages     = {459-468},
  doi       = {10.1109/CVPR42600.2020.00054}
}

@article{Wang2023bifuse2,
  author    = {Fu{-}En Wang and
               Yu{-}Hsuan Yeh and
               Yi{-}Hsuan Tsai and
               Wei{-}Chen Chiu and
               Min Sun},
  title     = {BiFuse++: Self-Supervised and Efficient Bi-Projection Fusion for 360{\textdegree}
               Depth Estimation},
  journal   = {{IEEE} Transactions on Pattern Analysis and Machine Intelligence},
  volume    = {45},
  number    = {5},
  pages     = {5448--5460},
  year      = {2023},
  doi       = {10.1109/TPAMI.2022.3203516}
}

@inproceedings{PanoSUNCG,
  title        = {Self-supervised Learning of Depth and Camera Motion from 360$^\circ$ Videos},
  author       = {Wang, Fu-En and Hu, Hou-Ning and Cheng, Hsien-Tzu and Lin, Juan-Ting and Yang, Shang-Ta and Shih, Meng-Li and Chu, Hung-Kuo and Sun, Min},
  booktitle    = {Asian Conference on Computer Vision, {ACCV}},
  pages        = {53--68},
  year         = {2019},
  organization = {Springer}
}

@inproceedings{360SD-Net,
  author    = {Wang, N. H. and Solarte, B. and Tsai, Y. H. and Chiu, W. C. and Sun, M.},
  title     = {360SD-Net: 360$^\circ$ Stereo Depth Estimation with Learnable Cost Volume},
  booktitle = {2020 IEEE International Conference on Robotics and Automation, {ICRA}},
  pages     = {582-588},
  isbn      = {2577-087X},
  doi       = {10.1109/ICRA40945.2020.9196975},
  year      = {2020}
}

@inproceedings{Omnimvs,
  author    = {Won, Changhee and Ryu, Jongbin and Lim, Jongwoo},
  title     = {Omnimvs: End-to-end learning for omnidirectional stereo matching},
  booktitle = {Proceedings of the IEEE/CVF International Conference on Computer Vision, {CVPR}},
  pages     = {8987-8996},
  year      = {2019}
}

@inproceedings{SweepNet,
  author    = {Won, C. and Ryu, J. and Lim, J.},
  title     = {SweepNet: Wide-baseline Omnidirectional Depth Estimation},
  booktitle = {2019 International Conference on Robotics and Automation, {ICRA}},
  pages     = {6073-6079},
  isbn      = {2577-087X},
  doi       = {10.1109/ICRA.2019.8793823},
  year      = {2019}
}

@article{Omnimvs_J,
  author   = {Won, Changhee and Ryu, Jongbin and Lim, Jongwoo},
  journal  = {IEEE Transactions on Pattern Analysis and Machine Intelligence},
  title    = {End-to-End Learning for Omnidirectional Stereo Matching With Uncertainty Prior},
  year     = {2021},
  volume   = {43},
  number   = {11},
  pages    = {3850-3862},
  doi      = {10.1109/TPAMI.2020.2992497}
}

@inproceedings{OmniVidar,
  author    = {Xie, Sheng and Wang, Daochuan and Liu, Yun-Hui},
  title     = {OmniVidar: Omnidirectional Depth Estimation From Multi-Fisheye Images},
  booktitle = {Proceedings of the IEEE/CVF Conference on Computer Vision and Pattern Recognition, {CVPR}},
  month     = {June},
  year      = {2023},
  pages     = {21529-21538}
}

@inproceedings{realtimeOmni,
  author    = {Meuleman, Andreas and Jang, Hyeonjoong and Jeon, Daniel S. and Kim, Min H.},
  booktitle = {2021 IEEE/CVF Conference on Computer Vision and Pattern Recognition, {CVPR}},
  title     = {Real-Time Sphere Sweeping Stereo from Multiview Fisheye Images},
  year      = {2021},
  volume    = {},
  number    = {},
  pages     = {11418-11427},
  doi       = {10.1109/CVPR46437.2021.01126}
}

@article{zbontar2016mccnn,
  author  = {Jure {\v{Z}}bontar and Yann LeCun},
  title   = {Stereo Matching by Training a Convolutional Neural Network to Compare Image Patches},
  journal = {Journal of Machine Learning Research},
  year    = {2016},
  volume  = {17},
  number  = {65},
  pages   = {1-32}
}

@inproceedings{3D60,
  author    = {Zioulis, N. and Karakottas, A. and Zarpalas, D. and Alvarez, F. and Daras, P.},
  title     = {Spherical View Synthesis for Self-Supervised 360° Depth Estimation},
  booktitle = {2019 International Conference on 3D Vision, {3DV}},
  pages     = {690-699},
  isbn      = {2475-7888},
  doi       = {10.1109/3DV.2019.00081},
  year      = {2019}
}

@article{Silog,
  title   = {Depth map prediction from a single image using a multi-scale deep network},
  author  = {Eigen, David and Puhrsch, Christian and Fergus, Rob},
  journal = {Advances in neural information processing systems},
  volume  = {27},
  year    = {2014}
}

@article{Jiang2021unifuse,
  author  = {Jiang, Hualie and Sheng, Zhe and Zhu, Siyu and Dong, Zilong and Huang, Rui},
  journal = {IEEE Robotics and Automation Letters},
  title   = {UniFuse: Unidirectional Fusion for 360° Panorama Depth Estimation},
  year    = {2021},
  volume  = {6},
  number  = {2},
  pages   = {1519-1526},
  doi     = {10.1109/LRA.2021.3058957}
}

@inproceedings{Delta123,
  author    = {Ladický, Lubor and Shi, Jianbo and Pollefeys, Marc},
  booktitle = {2014 IEEE Conference on Computer Vision and Pattern Recognition, {CVPR}},
  title     = {Pulling Things out of Perspective},
  year      = {2014},
  volume    = {},
  number    = {},
  pages     = {89-96},
  doi       = {10.1109/CVPR.2014.19}
}

@inproceedings{lipson2021raft,
  title     = {RAFT-Stereo: Multilevel Recurrent Field Transforms for Stereo Matching},
  author    = {Lipson, Lahav and Teed, Zachary and Deng, Jia},
  booktitle = {International Conference on 3D Vision, {3DV}},
  year      = {2021}
}

@inproceedings{Shen2021CFNet,
  author    = {Shen, Zhelun and Dai, Yuchao and Rao, Zhibo},
  title     = {CFNet: Cascade and Fused Cost Volume for Robust Stereo Matching},
  booktitle = {Proceedings of the IEEE/CVF Conference on Computer Vision and Pattern Recognition, {CVPR}},
  month     = {June},
  year      = {2021},
  pages     = {13906-13915}
}

@inproceedings{Li2022omnifusion,
  author    = {Yuyan Li and
               Yuliang Guo and
               Zhixin Yan and
               Xinyu Huang and
               Ye Duan and
               Liu Ren},
  title     = {OmniFusion: 360 Monocular Depth Estimation via Geometry-Aware Fusion},
  booktitle = {{IEEE/CVF} Conference on Computer Vision and Pattern Recognition, {CVPR}},
  pages     = {2791--2800},
  publisher = {{IEEE}},
  year      = {2022},
  doi       = {10.1109/CVPR52688.2022.00282}
}

@inproceedings{Feng2022segfuse,
  author    = {Qi Feng and
               Hubert P. H. Shum and
               Shigeo Morishima},
  title     = {360 Depth Estimation in the Wild - the Depth360 Dataset and the SegFuse
               Network},
  booktitle = {{IEEE} Conference on Virtual Reality and 3D User Interfaces, {VR}},
  pages     = {664--673},
  publisher = {{IEEE}},
  year      = {2022},
  doi       = {10.1109/VR51125.2022.00087}
}

@inproceedings{Li2022crestereo,
  author    = {Jiankun Li and
               Peisen Wang and
               Pengfei Xiong and
               Tao Cai and
               Ziwei Yan and
               Lei Yang and
               Jiangyu Liu and
               Haoqiang Fan and
               Shuaicheng Liu},
  title     = {Practical Stereo Matching via Cascaded Recurrent Network with Adaptive
               Correlation},
  booktitle = {{IEEE/CVF} Conference on Computer Vision and Pattern Recognition,
               {CVPR} },
  pages     = {16242--16251},
  publisher = {{IEEE}},
  year      = {2022},
  doi       = {10.1109/CVPR52688.2022.01578}
}

@inproceedings{Li2022mode,
  author    = {Li, Ming
               and Jin, Xueqian
               and Hu, Xuejiao
               and Dai, Jingzhao
               and Du, Sidan
               and Li, Yang},
  editor    = {Avidan, Shai
               and Brostow, Gabriel
               and Ciss{\'e}, Moustapha
               and Farinella, Giovanni Maria
               and Hassner, Tal},
  title     = {MODE: Multi-view Omnidirectional Depth Estimation with 360$^\circ$ Cameras},
  booktitle = {Proceedings of the 17th European Conference on Computer Vision, {ECCV}},
  year      = {2022},
  publisher = {Springer Nature Switzerland},
  address   = {Cham},
  pages     = {197--213},
  doi       = {10.1007/978-3-031-19827-4_12},
  isbn      = {978-3-031-19827-4}
}

@article{Su2023omni,
  author   = {Su, Xiaojie and Liu, Shimin and Li, Rui},
  journal  = {IEEE Transactions on Intelligent Transportation Systems},
  title    = {Omnidirectional Depth Estimation With Hierarchical Deep Network for Multi-Fisheye Navigation Systems},
  year     = {2023},
  volume   = {24},
  number   = {12},
  pages    = {13756-13767},
  doi      = {10.1109/TITS.2023.3294642}
}

@article{Zeng2024DMCA-Net,
  author   = {Zeng, Kai and Zhang, Hui and Wang, Wei and Wang, Yaonan and Mao, Jianxu},
  journal  = {IEEE Transactions on Circuits and Systems for Video Technology},
  title    = {Deep Stereo Network With MRF-Based Cost Aggregation},
  year     = {2024},
  volume   = {34},
  number   = {4},
  pages    = {2426-2438},
  doi      = {10.1109/TCSVT.2023.3312153}
}

@inproceedings{Chiu2023360MVSNet,
  author    = {Chiu, Ching-Ya and Wu, Yu-Ting and Shen, I-Chao and Chuang, Yung-Yu},
  booktitle = {2023 IEEE/CVF Winter Conference on Applications of Computer Vision, {WACV}},
  title     = {360MVSNet: Deep Multi-view Stereo Network with 360° Images for Indoor Scene Reconstruction},
  year      = {2023},
  volume    = {},
  number    = {},
  pages     = {3056-3065},
  doi       = {10.1109/WACV56688.2023.00307}
}

@article{Jiang2024Romnistereo,
  author   = {Jiang, Hualie and Xu, Rui and Tan, Minglang and Jiang, Wenjie},
  journal  = {IEEE Robotics and Automation Letters},
  title    = {RomniStereo: Recurrent Omnidirectional Stereo Matching},
  year     = {2024},
  volume   = {9},
  number   = {3},
  pages    = {2511-2518},
  doi      = {10.1109/LRA.2024.3357315}
}

@article{Wang2024Casomnimvs,
  author         = {Wang, Pinzhi and Li, Ming and Cao, Jinghao and Du, Sidan and Li, Yang},
  title          = {CasOmniMVS: Cascade Omnidirectional Depth Estimation with Dynamic Spherical Sweeping},
  journal        = {Applied Sciences},
  volume         = {14},
  year           = {2024},
  number         = {2},
  article-number = {517},
  issn           = {2076-3417},
  doi            = {10.3390/app14020517}
}

@inproceedings{Chen2023UnOmnimvs,
  author    = {Chen, Zisong and Lin, Chunyu and Nie, Lang and Liao, Kang and Zhao, Yao},
  booktitle = {2023 IEEE/RSJ International Conference on Intelligent Robots and Systems, {IROS}},
  title     = {Unsupervised OmniMVS: Efficient Omnidirectional Depth Inference via Establishing Pseudo-Stereo Supervision},
  year      = {2023},
  volume    = {},
  number    = {},
  pages     = {10873-10879},
  doi       = {10.1109/IROS55552.2023.10342332}
}

@inproceedings{Komatsu2020crown360,
  author    = {Ren Komatsu and
               Hiromitsu Fujii and
               Yusuke Tamura and
               Atsushi Yamashita and
               Hajime Asama},
  title     = {360{\textdegree} Depth Estimation from Multiple Fisheye Images with
               Origami Crown Representation of Icosahedron},
  booktitle = {Proceedings of the {IEEE/RSJ} International Conference on Intelligent Robots and Systems,
               {IROS}},
  pages     = {10092--10099},
  year      = {2020},
  doi       = {10.1109/IROS45743.2020.9340981}
}

@inproceedings{Wang2023FastOmniMVS,
  author    = {Wang, Yushen and Yang, Yang and Deng, Jiaxi and Meng, Haitao and Chen, Gang},
  booktitle = {2023 IEEE 29th International Conference on Parallel and Distributed Systems, {ICPADS}},
  title     = {FastOmniMVS: Real-time Omnidirectional Depth Estimation from Multiview Fisheye Images},
  year      = {2023},
  volume    = {},
  number    = {},
  pages     = {1311-1318},
  doi       = {10.1109/ICPADS60453.2023.00188}
}

@inproceedings{Girshick2015smoothl1,
  author    = {Ross B. Girshick},
  title     = {Fast {R-CNN}},
  booktitle = {Proceedings of the {IEEE} International Conference on Computer Vision, {ICCV}},
  pages     = {1440--1448},
  year      = {2015},
  url       = {https://doi.org/10.1109/ICCV.2015.169},
  doi       = {10.1109/ICCV.2015.169},
  bibsource = {dblp computer science bibliography, https://dblp.org}
}

@misc{Li2024robustmode,
  title         = {Robust and Flexible Omnidirectional Depth Estimation with Multiple 360$^\circ$ Cameras},
  author        = {Ming Li and Xueqian Jin and Xuejiao Hu and Jinghao Cao and Sidan Du and Yang Li},
  year          = {2024},
  eprint        = {2409.14766},
  archiveprefix = {arXiv},
  primaryclass  = {cs.CV},
  url           = {https://arxiv.org/abs/2409.14766}
}

@inproceedings{leeSemiSupervised2022,
  title     = {Semi-{{Supervised}} 360{$^\circ$} {{Depth Estimation}} from {{Multiple Fisheye Cameras}} with {{Pixel-Level Selective Loss}}},
  booktitle = {{{ICASSP}} 2022 - 2022 {{IEEE International Conference}} on {{Acoustics}}, {{Speech}} and {{Signal Processing}} ({{ICASSP}})},
  author    = {Lee, Jaewoo and Park, Daeul and Lee, Dongwook and Ji, Daehyun},
  year      = {2022},
  month     = may,
  pages     = {2290--2294},
  issn      = {2379-190X},
  doi       = {10.1109/ICASSP43922.2022.9746232}
}

@inproceedings{chenSOmniMVS2023,
  title      = {S-{{OmniMVS}}: {{Incorporating Sphere Geometry}} into {{Omnidirectional Stereo Matching}}},
  shorttitle = {S-{{OmniMVS}}},
  booktitle  = {Proceedings of the 31st {{ACM International Conference}} on {{Multimedia}}},
  author     = {Chen, Zisong and Lin, Chunyu and Nie, Lang and Shen, Zhijie and Liao, Kang and Cao, Yuanzhouhan and Zhao, Yao},
  year       = {2023},
  month      = oct,
  series     = {{{MM}} '23},
  pages      = {1495--1503},
  publisher  = {Association for Computing Machinery},
  address    = {New York, NY, USA},
  doi        = {10.1145/3581783.3612381}
}

@inproceedings{Deng2025OmniStereo,
  author    = {Deng, Jiaxi and Wang, Yushen and Meng, Haitao and Hou, Zuoxun and Chang, Yi and Chen, Gang},
  title     = {OmniStereo: Real-time Omnidireactional Depth Estimation with Multiview Fisheye Cameras},
  booktitle = {Proceedings of the Computer Vision and Pattern Recognition Conference (CVPR)},
  month     = {June},
  year      = {2025},
  pages     = {1003-1012}
}



\end{document}